\documentclass{article}

\usepackage{arxiv}

\usepackage[utf8]{inputenc} 
\usepackage[T1]{fontenc}    
\usepackage{url}            
\usepackage{booktabs}       
\usepackage{amsfonts}       
\usepackage{nicefrac}       
\usepackage{microtype}      
\usepackage{lipsum}		
\usepackage{graphicx}
\usepackage{natbib}
\usepackage{doi}

\usepackage{graphicx}
\usepackage{amsmath}
\usepackage{amssymb}

\usepackage{xcolor}

\usepackage[capitalize]{cleveref}
\crefname{section}{Sec.}{Secs.}
\Crefname{section}{Section}{Sections}
\crefname{table}{Tab.}{Tabs.}
\Crefname{table}{Table}{Tables}
\crefname{equation}{Equation}{Equations}

\newcommand{\secref}[1]{Section~\ref{#1}}
\newcommand{\figref}[1]{Figure~\ref{#1}}
\newcommand{\tabref}[1]{Table~\ref{#1}}
\renewcommand{\eqref}[1]{Equation~\ref{#1}}
\usepackage{tablefootnote}
\newcommand{\comment}[1]{}

\title{Generalized Pose Embeddings for Training In-the-Wild via Analysis-by-Synthesis
}

\date{October 15, 2022}	

\author{Dominik Borer \\
	ETH Zurich\\
	Disney Research | Studios\\
	\And
	Jakob Buhmann\\
	Disney Research | Studios\\
	\And
	Martin Guay\\
 Disney Research | Studios\\
}

\renewcommand{\headeright}{}
\renewcommand{\undertitle}{}
\renewcommand{\shorttitle}{\textit{Generalized Pose Embeddings for Training In-the-Wild via Analysis-by-Synthesis}}

\begin{document}
\maketitle

\newcommand{\FigOverview}{
\begin{figure*}[!h]
	\centering
	\includegraphics[width=1\linewidth]{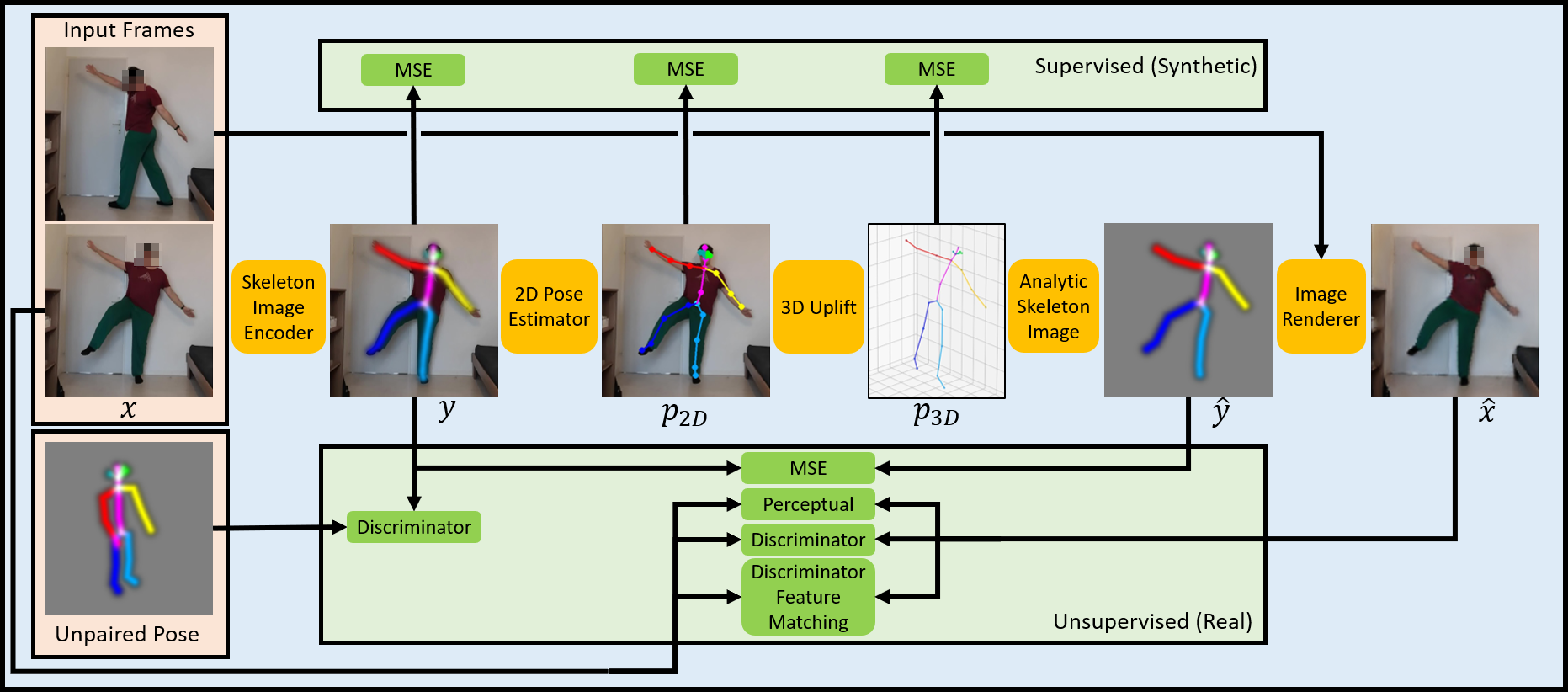}
	\caption{Overview of the model and training procedure. There are 5 main components (orange). First, the input image $x$ is mapped to a skeleton image representation $y$. From this the 2D pose $p_{2D}$ is estimated, which is then uplifted to 3D joint positions and orientations~$p_{3D}$. The reprojected coordinates are then used to analytically create a skeleton image~$\hat{y}$, from which the input image is reconstructed~$\hat{x}$. To train the model we use a mixture of synthetic and real data to optimize several objectives (green).}
	\label{fig:Overview}
\end{figure*}
}

\newcommand{\FigSyntheticData}{
\begin{figure}[!h]
	\centering
	\includegraphics[width=1\linewidth]{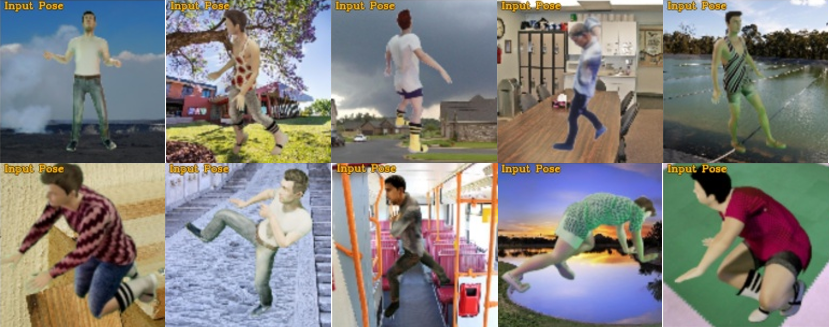}
	\caption{The synthetically generated data contains a lot of variation in pose, appearance and background.}
	\label{fig:SyntheticData}
\end{figure}
}

\newcommand{\FigRealData}{
\begin{figure}[!h]
	\centering
	\includegraphics[width=1\linewidth]{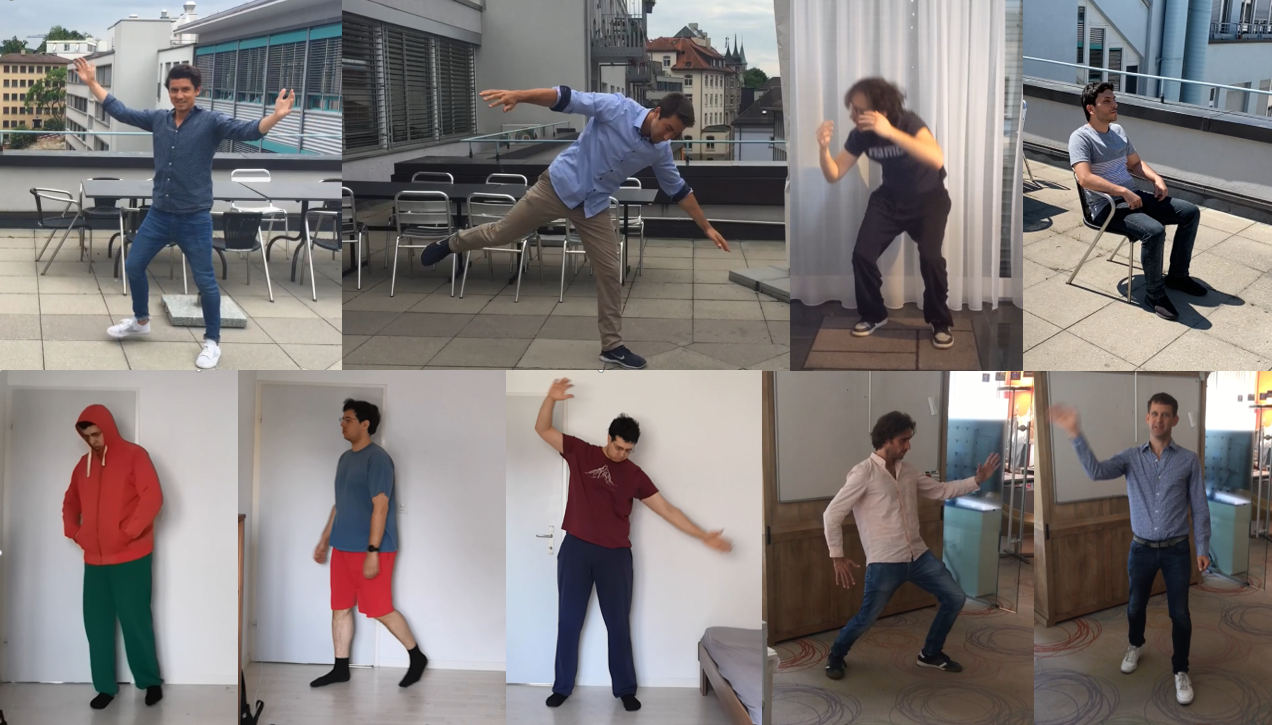}
	\caption{We use in-the-wild videos of a variety of different people performing various motions for the unlabelled, real data.}
	\label{fig:RealData}
\end{figure}
}

\newcommand{\FigSynthAndRealData}{
\begin{figure}[t]
	\centering
	\includegraphics[width=1\linewidth]{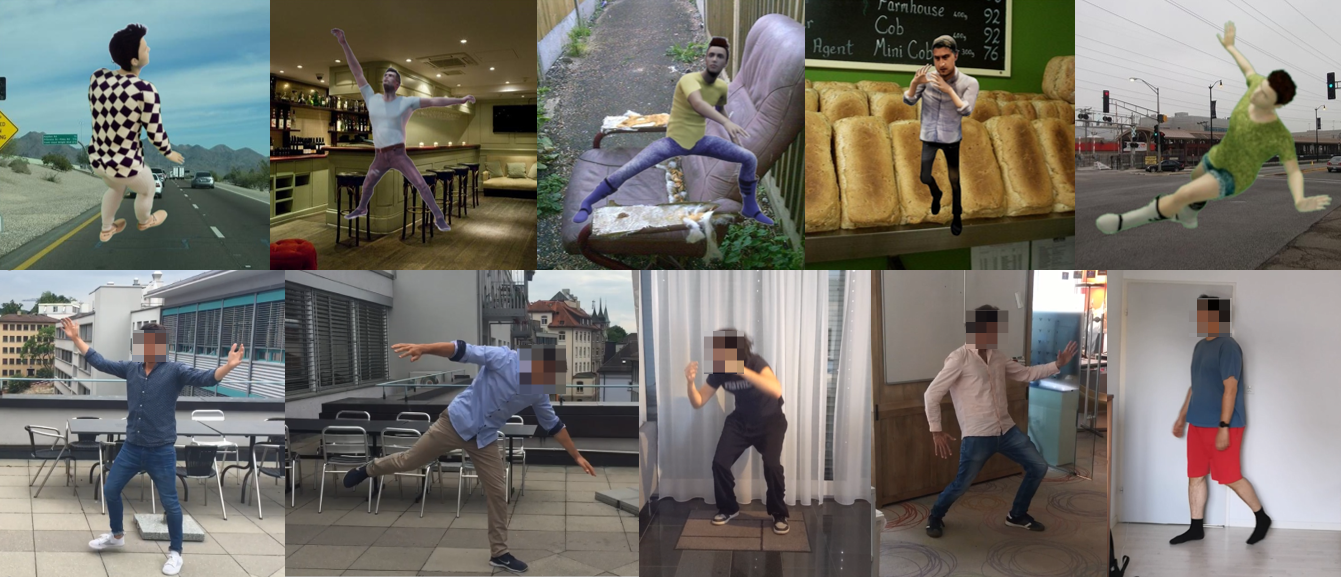}
	\caption{Samples from the training dataset. Top: Synthetically generated data. Bottom: Unlabelled, real, in-the-wild videos. The synthetic data contains a lot of variation in pose, appearance and background and the real data covers a variety of different people performing various motions.}
	\label{fig:SynthAndRealData}
\end{figure}
}

\newcommand{\FigSingleChannel}{
\begin{figure}[t]
	\centering
	\includegraphics[width=1\linewidth]{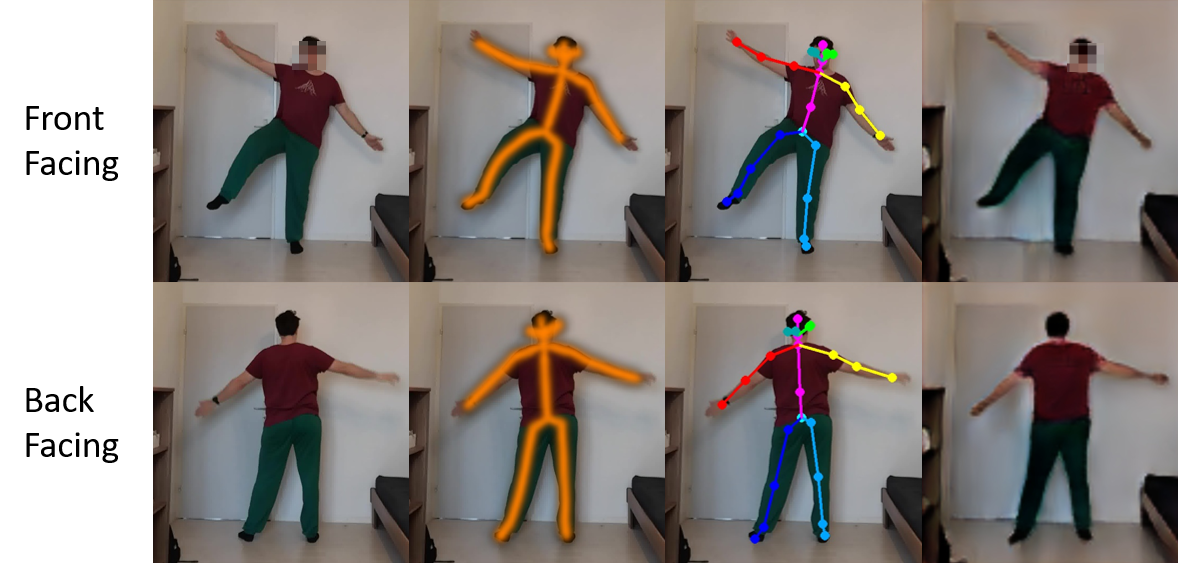}
	\caption{The single-channel skeleton image representation \cite{Jakab:2020:CVPR} suffers from ambiguities and fails to capture the body part semantics, causing flips in the predicted pose (e.g. red and blue should be the right arm and leg).}
	\label{fig:SingleChannel}
\end{figure}
}

\newcommand{\FigMultiChannel}{
\begin{figure}[t]
	\centering
	\includegraphics[width=1\linewidth]{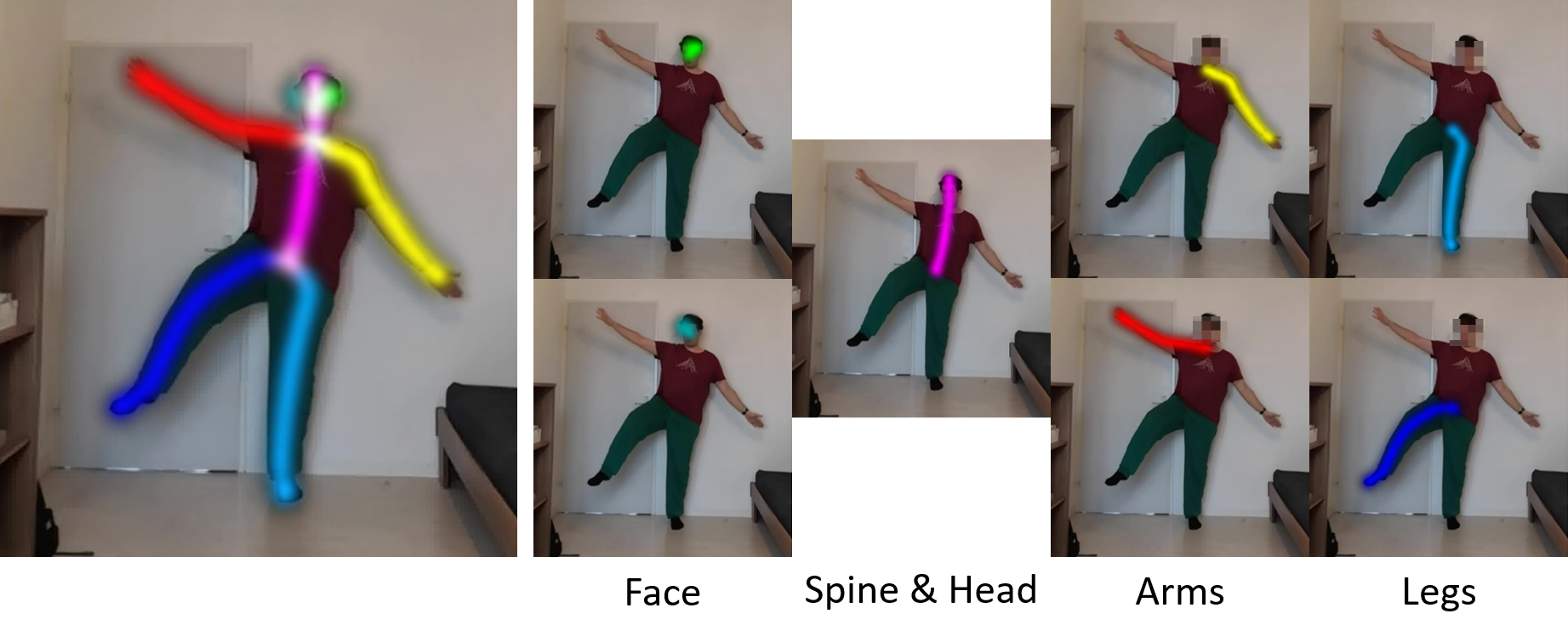}
	\caption{Our multi-channel skeleton image representation. Each channel (visualized with different colors) represents a semantically meaningful set of joints.}
	\label{fig:MultiChannel}
\end{figure}
}

\newcommand{\FigMultiChannelResult}{
\begin{figure}[!h]
	\centering
	\includegraphics[width=1\linewidth]{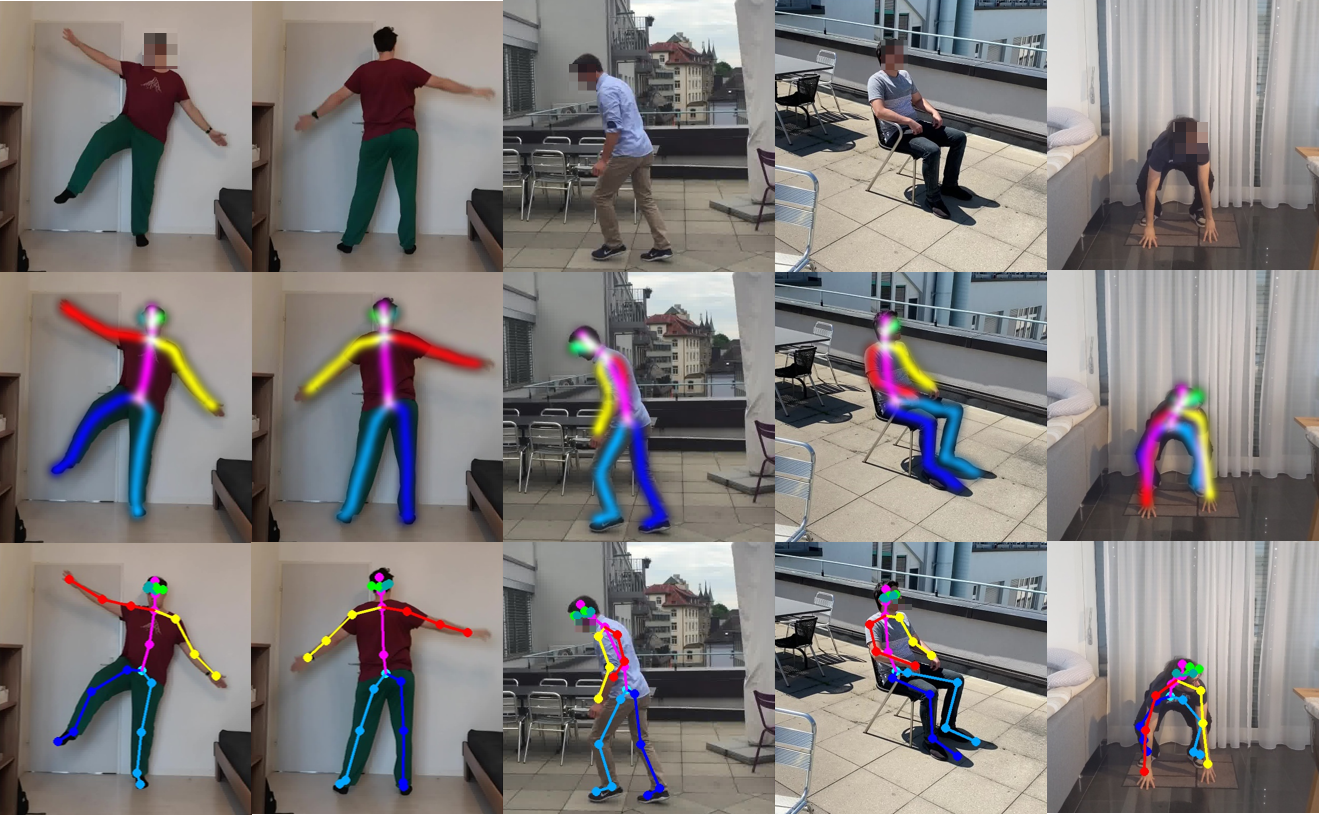}
	\caption{Predicted poses when using our multi-channel skeleton image. The predictions are accurate for a wide range of poses and do not suffer from left/right flips.}
	\label{fig:MultiChannelResult}
\end{figure}
}

\newcommand{\FigUplift}{
\begin{figure}[!h]
	\centering
	\includegraphics[width=1\linewidth]{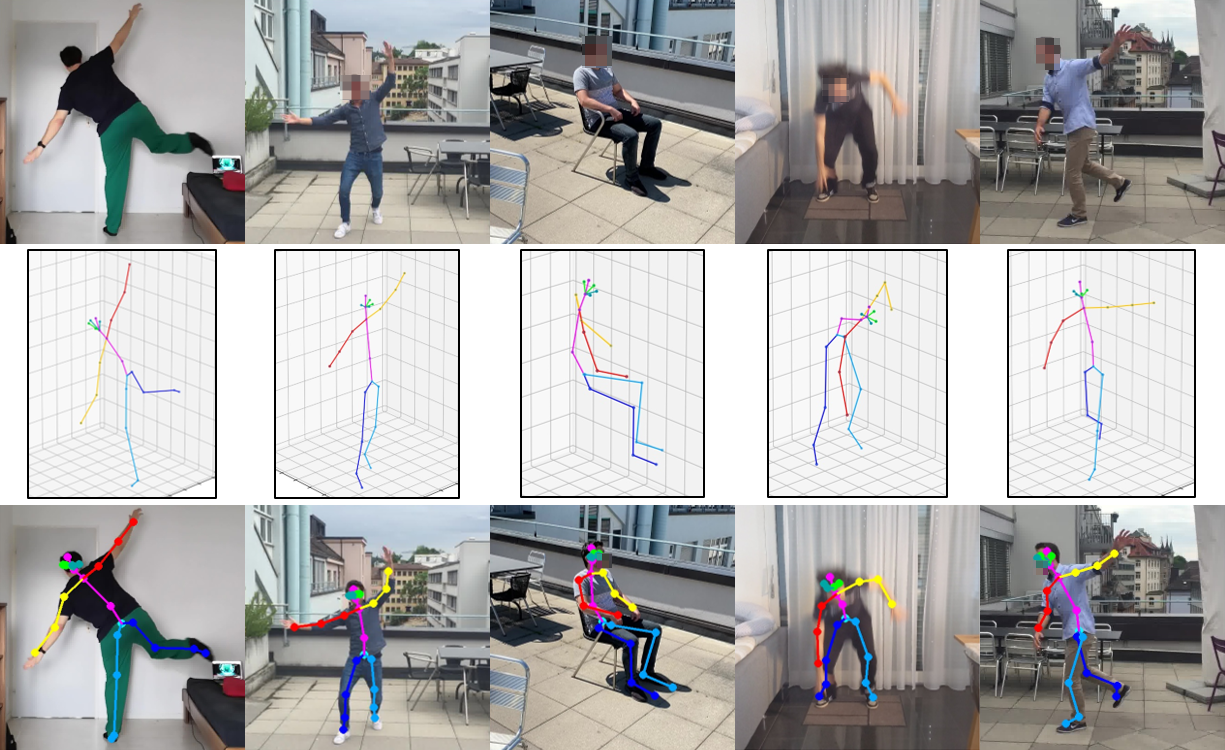}
	\caption{Uplifted 3D poses. Due to the end-to-end training, the poses are expressive and the reprojection (last row) overlaps with the person in the input image.}
	\label{fig:Uplift}
\end{figure}
}

\newcommand{\FigHM}{
\begin{figure}[t]
	\centering
	\includegraphics[width=1\linewidth]{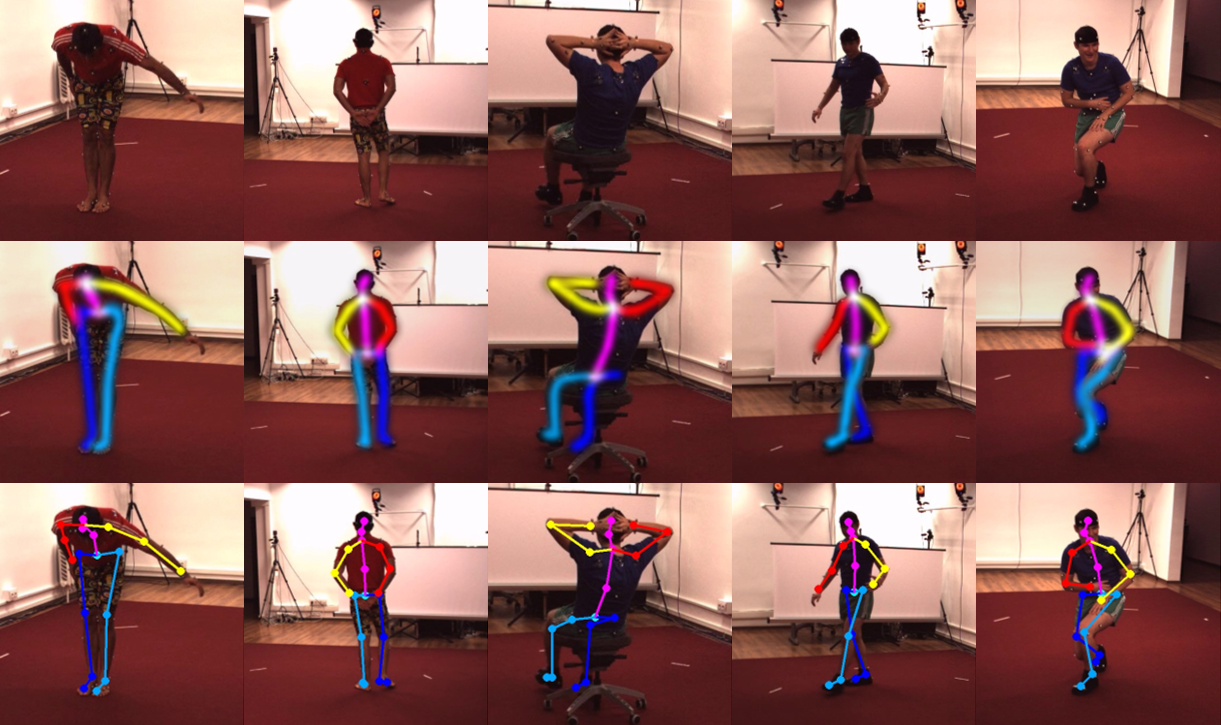}
	\caption{Results on the Human3.6M test set. The predictions are accurate and have no ambiguity issues. 
	}
	\label{fig:HM36}
\end{figure}
}

\newcommand{\FigInTheWild}{
\begin{figure*}[!h]
	\centering
	\includegraphics[width=1\linewidth]{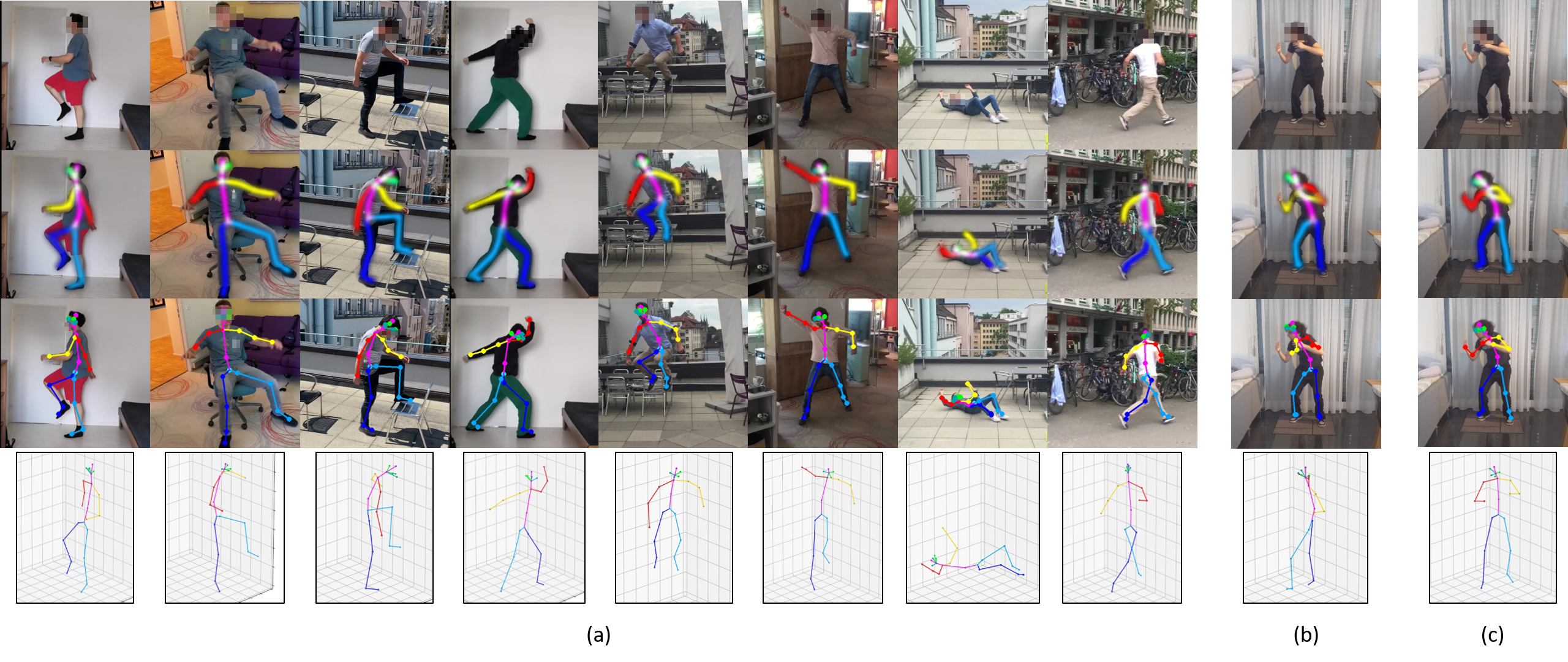}
	\caption{Predicted 2D and 3D poses for in-the-wild videos. Our approach can accurately estimate the pose of the person. For people used during training (a), it works very well. For a new, completely different person in a new environment (b), the model may have difficulties, but can quickly adapt through the unsupervised instance-specific refinement (c).}
	\label{fig:InTheWild}
\end{figure*}
}

\newcommand{\FigProportionsMismatch}{
\begin{figure}[!h]
	\centering
	\includegraphics[width=1\linewidth]{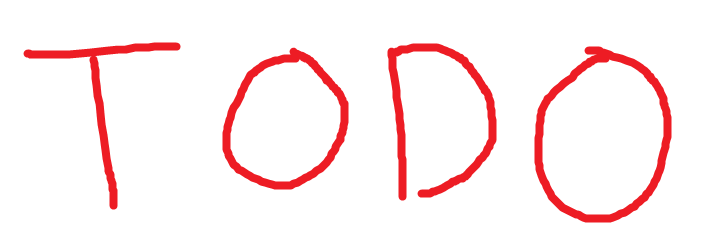}
	\caption{The data used to build the pose prior might have different limb proportions than the image data used for training. Training with such a prior without our limb length randomization causes incorrect predictions that do not overlap with the subject in the image.}
	\label{fig:ProportionsMismatch}
\end{figure}
}

\newcommand{\FigAnimals}{
\begin{figure}[!h]
	\centering
	\includegraphics[width=1\linewidth]{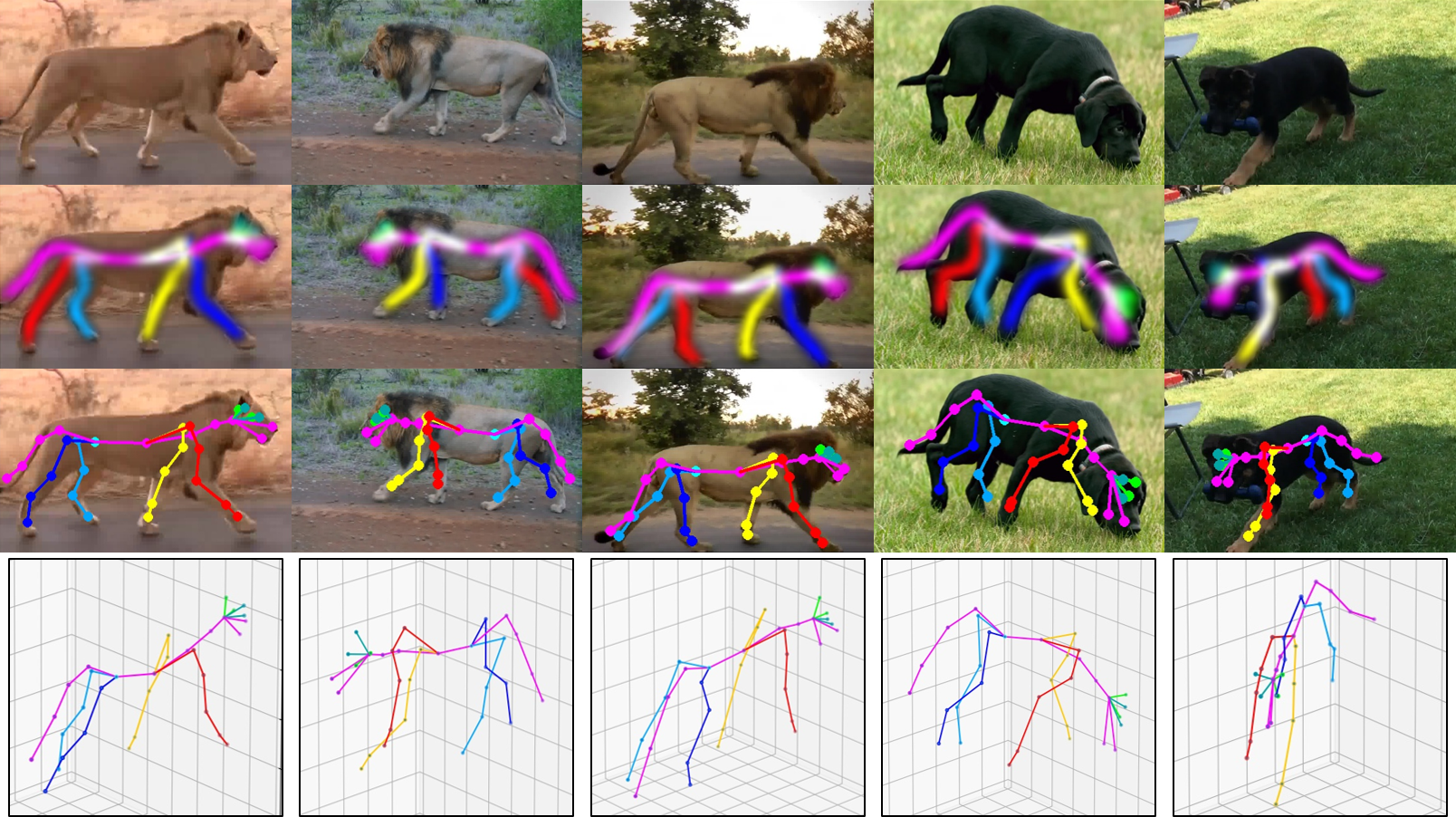}
	\caption{Estimated animal poses. Since the framework is generic it can be applied to other domains. We can successfully estimate poses for lions and dogs, that qualitatively outperform previous approaches such as \cite{Borer21:CatsAndDogs}.}
	\label{fig:Lion}
\end{figure}
}

\newcommand{\TableCombined}{
\begin{table}[t]
    \centering
    \small
    \begin{tabular}{ c|c|c|c } 
        \toprule
        & \cite{Jakab:2020:CVPR} & Ours & Ours Refined \\
        \midrule
        direct & 11.26 & 8.53 & \textbf{3.44} \\
        discuss & 13.72 & 12.25 & \textbf{5.98} \\
        eat & 12.02 & 10.43 & \textbf{6.36} \\
        greet & 11.85 & 11.47 & \textbf{4.56} \\
        phone & 14.42 & 16.97 & \textbf{9.33} \\
        pose & 10.39 & 16.82 & \textbf{2.80} \\
        purchases & 12.90 & 17.28 & \textbf{4.84} \\
        sit & 17.01 & 35.72 & \textbf{16.45} \\
        sit down & \textbf{25.71} & 64.30 & 28.66 \\
        smoke & 14.35 & 14.61 & \textbf{8.66} \\
        take photo & 18.67 & 13.09 & \textbf{9.41} \\
        wait & 11.40 & 15.43 & \textbf{4.45} \\
        walk & 11.85 & 11.08 & \textbf{3.26} \\
        walk dog & 19.42 & 10.71 & \textbf{5.07} \\
        walk together & 11.90 & 10.65 & \textbf{2.97} \\
        \midrule
        all & 14.46 & 17.96 & \textbf{7.75} \\
        \bottomrule
    \end{tabular}
    \caption{Comparison of the MSE in pixel to prior work on the Human3.6M test set. After training on real data (\emph{Ours}) we get better or similar scores and after refining (\emph{Ours Refined}) we outperform the baseline by a large margin. Note that \cite{Jakab:2020:CVPR} ignores flips by taking the smaller error of the current and flipped predictions,
    whereas we do not and hence the comparison is not entirely fair (additional comparisons in supplemental material).
    } 
    \label{tab:Human3.6M}
\end{table}
}

\newcommand{\TableSeparate}{
\begin{table}[h]
    \centering
    \begin{tabular}{ c|cc|cc } 
        \toprule
        & \multicolumn{2}{c|}{Ours} & \multicolumn{2}{c}{Ours Refined} \\
        & S9 & S11 & S9 & S11 \\
        \midrule
        direct & 17.39 & 2.53 & 3.78 & 2.01 \\
        discuss & 19.91 & 4.48 & 7.87 & 3.16 \\
        eat & 8.73 & 5.42 & 4.17 & 5.65 \\
        greet & 11.96 & 4.73 & 4.43 & 4.51 \\
        phone & 23.72 & 8.86 & 8.01 & 7.93 \\
        pose & 24.78 & 2.99 & 3.11 & 2.07 \\
        purchases & 28.31 & 6.68 & 5.79 & 4.31 \\
        sit & 119.55 & 23.08 & 59.50 & 21.63 \\
        sit down & 103.02 & 33.76 & 39.37 & 31.31 \\
        smoke & 17.82 & 9.24 & 6.27 & 8.13 \\
        take photo & 16.42 & 10.55 & 6.84 & 8.61 \\
        wait & 24.99 & 4.44 & 6.28 & 2.83 \\
        walk & 17.44 & 2.82 & 3.07 & 2.36 \\
        walk dog & 17.61 & 6.32 & 6.06 & 4.42 \\
        walk together & 15.35 & 4.03 & 2.60 & 2.39 \\
        \midrule
        all & 31.13 & 8.66 & 11.14 & 7.42 \\
        \bottomrule
    \end{tabular}
    \caption{MSE per pixel on the Human3.6M test set for subject 9 and 11 separately.
     The performance difference before refinement is due to the similarity between the subject 11 and the training subjects.}
    \label{tab:Human3.6MSeparate}
\end{table}
}

\newcommand{\TableAll}{
\begin{table*}[!htb]
    \centering
    \small
    \begin{tabular}{ c||c|c|c||cc|cc } 
        \toprule
        & \cite{Jakab:2020:CVPR} & Ours & Ours Refined & \multicolumn{2}{c|}{Ours} & \multicolumn{2}{c}{Ours Refined} \\
        & & & & S9 & S11 & S9 & S11 \\
        \midrule
        direct & 11.26 & 9.96 & \textbf{2.89} & 17.39 & 2.53 & 3.78 & 2.01 \\
        discuss & 13.72 & 12.20 & \textbf{5.51} & 19.91 & 4.48 & 7.87 & 3.16 \\
        eat & 12.02 & 7.08 & \textbf{4.91} & 8.73 & 5.42 & 4.17 & 5.65 \\
        greet & 11.85 & 8.35 & \textbf{4.47} & 11.96 & 4.73 & 4.43 & 4.51 \\
        phone & 14.42 & 16.29 & \textbf{7.97} & 23.72 & 8.86 & 8.01 & 7.93 \\
        pose & 10.39 & 13.89 & \textbf{2.59} & 24.78 & 2.99 & 3.11 & 2.07 \\
        purchases & 12.90 & 17.50 & \textbf{5.05} & 28.31 & 6.68 & 5.79 & 4.31 \\
        sit & \textbf{17.01} & 71.32 & 40.57 & 119.55 & 23.08 & 59.50 & 21.63 \\
        sit down & \textbf{25.71} & 68.39 & 35.34 & 103.02 & 33.76 & 39.37 & 31.31 \\
        smoke & 14.35 & 13.53 & \textbf{7.20} & 17.82 & 9.24 & 6.27 & 8.13 \\
        take photo & 18.67 & 13.49 & \textbf{7.72} & 16.42 & 10.55 & 6.84 & 8.61 \\
        wait & 11.40 & 14.72 & \textbf{4.56} & 24.99 & 4.44 & 6.28 & 2.83 \\
        walk & 11.85 & 10.13 & \textbf{2.71} & 17.44 & 2.82 & 3.07 & 2.36 \\
        walk dog & 19.42 & 11.97 & \textbf{5.24} & 17.61 & 6.32 & 6.06 & 4.42 \\
        walk together & 11.90 & 9.69 & \textbf{2.49} & 15.35 & 4.03 & 2.60 & 2.39 \\
        \midrule
        all & 14.46 & 19.90 & \textbf{9.28} & 31.13 & 8.66 & 11.14 & 7.42 \\
        \bottomrule
    \end{tabular}
    \caption{Comparison of the MSE per pixel on the Human3.6M test set to the prior work of \cite{Jakab:2020:CVPR}. After the unsupervised training on real data (\emph{Ours}) we outperform or achieve similar scores for most activities (higher errors for the sitting motions come from occluded keypoints, which we did not exclude in our metric). And after refining (\emph{Ours Refined}), we outperform the baseline by a large margin.
    }
    \label{tab:Human3.6MAll}
\end{table*}
}

\newcommand{\TableIgnoreFlip}{
\begin{table*}[t]
    \centering
    \small
    \begin{tabular}{ c||c||c|c|c||c|c|c||c|c } 
        \toprule
        (ignore flip) & \cite{Jakab:2020:CVPR} & \multicolumn{3}{c||}{Ours} & \multicolumn{3}{c||}{Ours Refined} & \multicolumn{2}{c}{Ours No-Synth (5-ch)} \\
        & 1-ch & 1-ch & 3-ch & 5-ch & 1-ch & 3-ch & 5-ch & Unrefined & Refined \\
        \midrule
        direct          	& 11.26	    & 6.38	    & 4.73	    & 4.83	    & 6.19	    & 4.26	    & \textbf{2.98}      & 35.51     & 33.33     \\
        discuss         	& 13.72	    & 9.52	    & 6.81	    & 7.18	    & 8.35	    & 7.21	    & \textbf{4.93}      & 49.06     & 43.10     \\
        eat             	& 12.02	    & 8.12	    & 6.23	    & 6.21	    & 7.59	    & 6.20	    & \textbf{5.97}      & 55.41     & 50.45     \\
        greet           	& 11.85	    & 7.91	    & 6.13	    & 6.11	    & 6.61	    & 5.58	    & \textbf{3.66}      & 47.67     & 43.21     \\
        phone           	& 14.42	    & 17.94	    & 10.70	    & 10.90	    & 16.39	    & 10.97	    & \textbf{8.17}      & 75.52     & 72.62     \\
        pose            	& 10.39	    & 5.48	    & 5.13	    & 6.53	    & 5.53	    & 4.59	    & \textbf{2.67}      & 37.29     & 33.87     \\
        purchases       	& 12.90	    & 20.81	    & 10.60	    & 10.44	    & 15.59	    & 8.47	    & \textbf{4.77}      & 66.81     & 51.62     \\
        sit             	& 17.01	    & 32.87	    & 22.30	    & 25.24	    & 28.01	    & 27.09	    & \textbf{13.86}     & 146.53    & 139.44    \\
        sit down	        & 25.71	    & 78.13	    & 37.18	    & 37.52	    & 73.32	    & 29.54	    & \textbf{22.65}     & 191.83    & 135.04    \\
        smoke           	& 14.35	    & 19.27	    & 10.29	    & 9.15	    & 17.19	    & 9.98	    & \textbf{7.19}      & 77.54     & 71.20     \\
        take photo	        & 18.67	    & 16.58	    & 9.07	    & 8.44	    & 15.73	    & 8.97	    & \textbf{7.72}      & 67.13     & 57.96     \\
        wait            	& 11.40	    & 12.00	    & 7.35	    & 6.53	    & 8.33	    & 5.96	    & \textbf{4.12}      & 44.18     & 39.36     \\
        walk            	& 11.85	    & 6.13	    & 4.52	    & 5.32	    & 5.90	    & 4.18	    & \textbf{3.09}      & 47.77     & 46.48     \\
        walk dog	        & 19.42	    & 11.66	    & 6.81	    & 6.80	    & 10.22	    & 6.56	    & \textbf{4.55}      & 52.00     & 43.38     \\
        walk together	    & 11.90	    & 7.31	    & 4.82	    & 4.65	    & 6.18	    & 3.84	    & \textbf{2.91}      & 49.45     & 44.80     \\
        \midrule
        all 	            & 14.46	    & 17.34	    & 10.18	    & 10.39	    & 15.41	    & 9.56	    & \textbf{6.62}      & 69.58     & 60.39     \\
        \bottomrule
    \end{tabular}
    \caption{Comparison of the MSE in pixel to prior work on the Human3.6M test set. We compare the performance of different configurations (\emph{1-}, \emph{2-} and \emph{3-channels}), the improvements through the refinement, as well as the importance of the synthetic supervision (\emph{No-Synth}). After training on real data (\emph{Ours}) we achieve better or similar scores and after refining (\emph{Ours Refined}) we outperform the baseline by a large margin. Configurations with more channels show to be superior over a single channel. Without synthetic supervision (\emph{Ours No-Synth}) the training fails to converge to a good solution, leading to poor accuracy. Note that as in \cite{Jakab:2020:CVPR} this metric ignores flips by taking the smaller error of the current and the flipped predictions. In \tabref{tab:Human3.6M:WithFlip} we compare the results where flips are considered.
    } 
    \label{tab:Human3.6M:IgnoreFlip}
\end{table*}
}

\newcommand{\TableConsiderFlip}{
\begin{table*}[t]
    \centering
    \small
    \begin{tabular}{ c||c||c|c|c||c|c|c||c|c } 
        \toprule
        (consider flip) & \cite{Jakab:2020:CVPR} & \multicolumn{3}{c||}{Ours} & \multicolumn{3}{c||}{Ours Refined} & \multicolumn{2}{c}{Ours No-Synth (5-ch)} \\
        & 1-ch & 1-ch & 3-ch & 5-ch & 1-ch & 3-ch & 5-ch & Unrefined & Refined \\
        \midrule
        direct              & 11.26     & 30.69	    & 5.83	    & 8.53	    & 24.41	    & 4.94	    & \textbf{3.44}      & 73.74	    & 44.98     \\
        discuss             & 13.72     & 41.05	    & 9.16	    & 12.25	    & 32.52	    & 8.95	    & \textbf{5.98}      & 90.11	    & 55.64     \\
        eat                 & 12.02     & 52.76	    & 7.72	    & 10.43	    & 43.19	    & 6.89	    & \textbf{6.36}      & 80.89	    & 60.90     \\
        greet               & 11.85     & 34.81	    & 7.82	    & 11.47	    & 27.69	    & 6.34	    & \textbf{4.56}      & 76.33	    & 52.36     \\
        phone               & 14.42     & 50.77	    & 12.75	    & 16.97	    & 41.75	    & 11.98	    & \textbf{9.33}      & 108.59	    & 93.08     \\
        pose                & 10.39     & 40.67	    & 7.36	    & 16.82	    & 31.92	    & 5.16	    & \textbf{2.80}      & 90.09	    & 44.07     \\
        purchases           & 12.90     & 89.88	    & 12.71	    & 17.28	    & 72.54	    & 8.72	    & \textbf{4.84}      & 99.13	    & 63.61     \\
        sit                 & 17.01     & 93.59	    & 34.19	    & 35.72	    & 82.45	    & 37.02	    & \textbf{16.45}     & 193.50	    & 191.91    \\
        sit down            & \textbf{25.71}     & 161.64	& 58.27	    & 64.30	    & 160.72	& 40.22	    & 28.66     & 305.54	    & 171.18    \\
        smoke               & 14.35     & 60.68	    & 15.02	    & 14.61	    & 50.48	    & 11.98	    & \textbf{8.66}      & 120.18	    & 97.89     \\
        take photo          & 18.67     & 70.82	    & 12.85	    & 13.09	    & 62.46	    & 10.23	    & \textbf{9.41}      & 108.12	    & 80.05     \\
        wait                & 11.40     & 47.06	    & 10.64	    & 15.43	    & 39.95	    & 6.53	    & \textbf{4.45}      & 85.68	    & 52.57     \\
        walk                & 11.85     & 35.43	    & 7.00	    & 11.08	    & 28.46	    & 4.45	    & \textbf{3.26}      & 81.67	    & 60.82     \\
        walk dog            & 19.42     & 40.08	    & 9.42	    & 10.71	    & 36.52	    & 7.72	    & \textbf{5.07}      & 87.84	    & 65.11     \\
        walk together       & 11.90     & 35.74	    & 7.31	    & 10.65	    & 29.28	    & 4.07	    & \textbf{2.97}      & 93.87	    & 54.46     \\
        \midrule
        all                 & 14.46     & 59.04	    & 14.54	    & 17.96	    & 50.96	    & 11.68	    & \textbf{7.75}      & 113.02	& 79.24     \\
        \bottomrule
    \end{tabular}
    \caption{Comparison of the MSE in pixel on the Human3.6M test set, where the metric does not ignore flips, in contrast to \cite{Jakab:2020:CVPR}. Compared to \tabref{tab:Human3.6M:IgnoreFlip} we observe a much higher error for the single-channel configuration (\emph{1-ch}), which shows that the single-channel skeleton representation is not expressive enough and suffers from many flips. For the multi-channel skeleton configurations (\emph{3-ch}, \emph{5-ch}) the difference is much smaller, especially after the refinement, which shows the effectiveness of the multi-channel representation.
    } 
    \label{tab:Human3.6M:WithFlip}
\end{table*}
}

\newcommand{\TableIgnoreFlipMedian}{
\begin{table*}[t]
    \centering
    \small
    \begin{tabular}{ c||c|c|c||c|c|c||c|c } 
        \toprule
        (ignore flip) & \multicolumn{3}{c||}{Ours} & \multicolumn{3}{c||}{Ours Refined} & \multicolumn{2}{c}{Ours No-Synth (5-ch)} \\
        (median) & 1-ch & 3-ch & 5-ch & 1-ch & 3-ch & 5-ch & Unrefined & Refined \\
        \midrule
        direct          & 3.58      & 2.96      & 3.19      & 3.56      & 2.90      & 2.24      & 30.04     & 28.48     \\
        discuss         & 4.88      & 3.68      & 3.76      & 4.38      & 3.38      & 2.63      & 35.16     & 32.22     \\
        eat             & 5.79      & 4.06      & 4.30      & 5.07      & 4.01      & 3.84      & 47.49     & 41.86     \\
        greet           & 4.17      & 3.33      & 3.72      & 3.76      & 3.26      & 2.54      & 37.48     & 33.21     \\
        phone           & 8.12      & 5.35      & 5.93      & 7.89      & 4.78      & 4.27      & 59.30     & 52.07     \\
        pose            & 3.26      & 3.02      & 3.38      & 2.72      & 2.94      & 2.00      & 31.56     & 28.16     \\
        purchases       & 5.41      & 3.90      & 4.29      & 5.41      & 3.73      & 2.99      & 38.95     & 34.45     \\
        sit             & 16.06     & 8.11      & 7.53      & 11.87     & 7.67      & 6.13      & 91.86     & 79.02     \\
        sit down	    & 60.47     & 22.26     & 22.57     & 58.83     & 19.52     & 14.79     & 165.91    & 108.09    \\
        smoke           & 8.27      & 5.52      & 5.02      & 7.03      & 4.94      & 4.00      & 56.67     & 48.11     \\
        take photo	    & 7.54      & 4.34      & 4.52      & 8.40      & 4.87      & 3.81      & 45.33     & 43.07     \\
        wait            & 4.78      & 3.65      & 3.91      & 4.49      & 3.17      & 2.58      & 34.22     & 31.33     \\
        walk            & 4.63      & 3.31      & 3.60      & 4.77      & 3.18      & 2.43      & 32.73     & 31.75     \\
        walk dog	    & 7.00      & 4.92      & 4.79      & 7.29      & 4.74      & 3.47      & 42.94     & 38.10     \\
        walk together	& 3.91      & 2.97      & 2.91      & 3.55      & 2.65      & 2.20      & 34.56     & 31.13     \\
        \midrule
        all 	        & 9.86      & 5.43      & 5.56      & 9.27      & 5.05      & 4.00      & 52.28     & 44.07     \\        
        \bottomrule
    \end{tabular}
    \caption{Comparison of the median of the squared errors in pixel on the Human3.6M test set. We compare the performance of different configurations (\emph{1-}, \emph{2-} and \emph{3-channels}), the improvements through the refinement (\emph{Refined}), as well as the importance of the synthetic supervision (\emph{No-Synth}). Configurations with more channels show to be superior over a single channel. Without synthetic supervision (\emph{Ours No-Synth}) the training fails to converge to a good solution, leading to poor accuracy. Note that as in \cite{Jakab:2020:CVPR} this metric ignores flips by taking the smaller error of the current and the flipped predictions. In \tabref{tab:Human3.6M:WithFlipMedian} we compare the results where flips are considered.
    } 
    \label{tab:Human3.6M:IgnoreFlipMedian}
\end{table*}
}

\newcommand{\TableConsiderFlipMedian}{
\begin{table*}[t]
    \centering
    \small
    \begin{tabular}{ c||c|c|c||c|c|c||c|c } 
        \toprule
        (consider flip) & \multicolumn{3}{c||}{Ours} & \multicolumn{3}{c||}{Ours Refined} & \multicolumn{2}{c}{Ours No-Synth (5-ch)} \\
        (median) & 1-ch & 3-ch & 5-ch & 1-ch & 3-ch & 5-ch & Unrefined & Refined \\
        \midrule
        direct          & 29.53     & 3.11      & 3.58      & 24.65     & 3.00      & 2.29      & 46.92     & 34.75     \\
        discuss         & 29.24     & 3.99      & 4.17      & 24.20     & 3.53      & 2.72      & 53.33     & 38.94     \\
        eat             & 29.10     & 4.15      & 4.39      & 28.26     & 4.04      & 3.88      & 63.40     & 47.64     \\
        greet           & 19.52     & 3.36      & 3.88      & 18.36     & 3.29      & 2.56      & 49.66     & 40.25     \\
        phone           & 40.14     & 5.76      & 6.97      & 32.84     & 4.97      & 4.45      & 83.41     & 65.66     \\
        pose            & 22.93     & 3.08      & 3.94      & 22.13     & 2.97      & 2.03      & 42.78     & 30.13     \\
        purchases       & 53.93     & 4.03      & 4.39      & 40.04     & 3.84      & 3.00      & 64.12     & 41.76     \\
        sit             & 73.74     & 8.52      & 7.99      & 60.67     & 8.27      & 6.18      & 123.41    & 102.73    \\
        sit down	    & 106.35    & 25.67     & 30.97     & 103.66    & 21.88     & 16.47     & 231.37    & 123.16    \\
        smoke           & 36.41     & 6.16      & 5.48      & 33.47     & 5.26      & 4.28      & 91.07     & 66.90     \\
        take photo	    & 29.61     & 4.47      & 4.87      & 30.39     & 5.25      & 4.04      & 76.35     & 61.50     \\
        wait            & 23.67     & 4.16      & 5.20      & 22.35     & 3.43      & 2.79      & 58.17     & 40.13     \\
        walk            & 17.94     & 3.42      & 3.92      & 16.97     & 3.24      & 2.48      & 59.86     & 37.97     \\
        walk dog	    & 25.01     & 5.32      & 5.47      & 22.99     & 5.02      & 3.64      & 68.05     & 49.59     \\
        walk together	& 18.81     & 3.01      & 3.09      & 16.94     & 2.68      & 2.23      & 58.77     & 34.21     \\
        \midrule
        all 	        & 37.06     & 5.88      & 6.55      & 33.19     & 5.38      & 4.20      & 78.04     & 54.35     \\
        \bottomrule
    \end{tabular}
    \caption{Comparison of the median of the squared errors in pixel on the Human3.6M test set, where the metric does not ignore flips, in contrast to \cite{Jakab:2020:CVPR}. Compared to \tabref{tab:Human3.6M:IgnoreFlipMedian} we observe a much higher error for the single-channel configuration (\emph{1-ch}), which shows that the single-channel skeleton representation is not expressive enough and suffers from many flips. For the multi-channel skeleton image configurations (\emph{3-ch}, \emph{5-ch}) the difference is much smaller, especially after the refinement, which shows the effectiveness of the multi-channel representation.
    } 
    \label{tab:Human3.6M:WithFlipMedian}
\end{table*}
}

\begin{abstract}

Modern pose estimation models are trained on large, manually-labelled datasets which are costly and may not cover the full extent of human poses and appearances in the real world.
With advances in neural rendering, analysis-by-synthesis and the ability to not only predict, but also render the pose, is becoming an appealing framework, which could alleviate the need for large scale manual labelling efforts.
While recent work have shown the feasibility of this approach, the predictions admit many flips due to a simplistic intermediate skeleton representation, resulting in low precision and inhibiting the acquisition of any downstream knowledge such as three-dimensional positioning.
We solve this problem with a more expressive intermediate skeleton representation capable of capturing the semantics of the pose (left and right), which significantly reduces flips. To successfully train this new representation, we extend the analysis-by-synthesis framework with a training protocol based on synthetic data. We show that our representation results in less flips and more accurate predictions.
Our approach outperforms previous models trained with analysis-by-synthesis on standard benchmarks.

\end{abstract}

\FigOverview

\section{Introduction}

Building datasets for pose estimation of humans and animals is costly and remains challenging in terms of capturing the full diversity in pose and appearance of the real world. Some works seek to utilize synthetic humans but remain at a gap from real world imagery. This inhibits the ability of labelled datasets---both synthetic and real---to generalize and motivates the need for a class of methods that can learn from un-labelled video data in-the-wild.

Analysis-by-synthesis is the idea of including a depiction model in order to define a loss or cost back to the input image---enabling a training signal directly from unlabelled images.   
Such an approach, within a deep learning human pose estimation setting has been demonstrated in \cite{Jakab:2020:CVPR}, where they use an intermediate pictorial skeleton representation of the pose. While their representation of a pose can be trained in an unsupervised way, it consists of only a single channel ($\mathbb{R}^{1 \times W \times H}$) and does not represent the semantics of body parts---leading to significant flips between left and right, front and back in both the pose predictions and renderings of the person, as can be seen in \figref{fig:SingleChannel}. 

We solve this problem with a new multi-channel pose representation. However, we found this higher capacity model sensitive to divergence during training. To address this problem, we introduce synthetic humans to pre-train our pose and rendering model, providing a better conditioning for further fine-tuning using analysis-by-synthesis in the wild. Another way to look at our approach is to consider analysis-by-synthesis a tool for bridging the reality gap of models trained with synthetic data.

On the standard benchmark Human3.6M~\cite{h36m:pami}, we outperform the prior work of Jakab~\cite{Jakab:2020:CVPR} with an MSE of 10.39, compared to the MSE of 14.46 for the baseline.
We also gathered specialized data of a target subject and measured whether fine-tuning on this data would improve accuracy at run-time. On the Human3.6M benchmark, this refinement step further reduces the MSE to 6.62, outperforming the baseline by a large margin. Additionally, we investigated our framework further to answer whether a 3D pose representation could be trained end-to-end. We also show that our approach can generalize to other skeletal structures such as animals, where we qualitatively improve the accuracy of the 3D pose predictions compared to the work of Borer~\cite{Borer21:CatsAndDogs}.

\section{Related Work} \label{sec:RelatedWork}

Our work sits at the intersection of supervised and un-supervised pose estimation, where neural rendering enables learning from un-labelled images, while supervision is provided by synthetic data, both to initialize our model, and to provide a noise-free pose prior. 


Fully supervised methods leverage large-scale datasets with 2D and/or 3D annotations such as MS~COCO \cite{Lin14}, Human3.6M \cite{h36m:pami}, CMU \cite{CMUmocap}, and MPII \cite{Andriluka:2014}. Such approaches usually do not need an additional pose prior, as it is already captured in the data. Approaches for 2D pose estimation use CNNs to regress keypoint confidence maps \cite{Wei2016,Newell2016}, which can be refined iteratively \cite{Cao2018}, or directly regress keypoint coordinates \cite{Toshev:2014}. For 3D pose estimation it is common to train a separate model to go from 2D keypoints to 3D joint locations and orientations \cite{Martinez17}, but others also directly regress to 3D poses from images \cite{Borer21:CatsAndDogs}. Because hand-labelling datasets is very expensive and motion capture datasets lack variation in the images, other works train from synthetically generated data \cite{Varol:2017}. Even though not as accurate as training on real data, this has proven to be practical for exposing DNNs to 3D information, as well as to domains such as animals, where annotated data is not available or very hard to acquire \cite{zuffi20163D,Zuffi:ICCV:2019,Borer21:CatsAndDogs}. 
In this work, we also rely on synthetic data, but combine it with unlabelled real data to address the reality gap.

\paragraph{Weak Supervision}
Methods with weak supervision assume that only some annotations are available. In \cite{hmrKanazawa17,Pavlakos:2018,deepcap:2020} a dense 3D human mesh is predicted from sparse 2D keypoint annotations, with a loss on the reprojected keypoints. To ensure a plausible 3D shape, the SMPL parametric human mesh model \cite{SMPL:2015} is used as a prior. This is combined with motion capture data for adversarial learning \cite{hmrKanazawa17,Yang:2018}, a loss on the rendered silhouette \cite{Pavlakos:2018}, or multi-view consistency \cite{deepcap:2020}.
Similarly, we also use motion capture data for an empirical pose prior, but we do not need a fully fledged 3D prior such as provided by the SMPL model.

\paragraph{Un-supervised Pose Estimation}
On the other hand, unsupervised methods do not need any annotations. Works such as \cite{Kanazawa:2016} learn to match pairs of images of an object, but do not learn geometric information like the pose. Landmarks as an explicit structural representation can be learned through analysis-by-synthesis, by reconstructing the input image conditioned on the landmarks \cite{Thewlis:2017:ICCV,Jakab2018,2018-cvpr-lmdis-rep,Lorenz:2019}. But these landmarks are not constrained to any semantics. Hence, additional paired supervision is required to map the landmarks to the desired semantics. Other approaches tackle this lack of semantics by fitting a template model to the landmarks \cite{Kundu:2020,Schmidtke:2021:CVPR}.
Directly using a pose prior inside the network, forces the landmarks to follow the desired pose representation without the need for a template model \cite{Jakab:2020:CVPR}. While this directly learns a pose structure, it remains limited to a single channel and admits significant left/right ambiguities. We  solve this problem in this work with a richer pose model that is directly learned and shows an increase in performance. 

Another line of work in unsupervised pose estimation relies on multi-view consistency \cite{Rhodin:2018,Chen:2019}. Given the predicted pose from one view, the image in a different view is reconstructed \cite{Rhodin:2018}. Or for monocular data the predicted 3D pose is projected to a different camera and uplifted again to compare to the original pose \cite{Chen:2019}. These training schemes are orthogonal to our work and could easily be integrated into our framework.

\paragraph{Adversarial Learning - Neural Rendering}
For the approach of analysis-by-synthesis to work, the original input image has to be reconstructed, which creates the training signal. This unsupervised training falls into the domain of neural rendering and generative imagery. To achieve this, adversarial learning \cite{Goodfellow:2014:GAN} is useful to train a deep generative network to translate images from one domain to another using paired \cite{pix2pix2017} or unpaired \cite{CycleGAN2017} samples, and is further improved to high resolution images in \cite{Wang:2018:pix2pixHD,park2019SPADE}. Similar ideas are also applied to articulated characters by translating an image of a skeleton \cite{Chan:2019:EverybodyDance,Aberman:2019} or a segmentation map \cite{Sarkar:2021:HumanGAN} to an image of a person or to directly map a 3D pose vector to a high resolution image \cite{Borer21:RigSpace}.
In our case, we use image-to-image translation networks to predict the skeleton image and to reconstruct the input image.


\FigSynthAndRealData

\section{Dataset} \label{sec:dataset}
To address the data annotation issue, we use a mixture of synthetic data as well as unlabelled real-world data. This way we do not rely on manual labelling, and use the synthetic data to pre-train our model, while images-in-the-wild are used to fine tune our network to real world scenarios.

To generate the synthetic data, we use a setup similar to \cite{Borer21:CatsAndDogs}. We use domain randomization \cite{Tobin2017} to create variations in appearance, shape and pose, as can be seen in the top row of \figref{fig:SynthAndRealData}. As shown in \cite{Borer21:CatsAndDogs}, a model trained on such data has the ability to detect poses in the real-world, but it remains at a gap from models trained on real-world data, which we provide in this work through analysis-by-synthesis.

Our unlabelled real-world data consists of videos in-the-wild comprising different people, performing various actions in different environments, as can be seen in the bottom row of \figref{fig:SynthAndRealData}. While we do not strictly require video data, we need at least two frames of the same person in the same environment. We employ a pose prior that is trained on an in-house 3D motion capture dataset similar to Human3.6M \cite{h36m:pami}. Using the synthetic data generation pipeline, we extend this pose data with additional surface keypoints, i.e.\ facial keypoints, to further help with the ambiguity issue.


\section{Method} \label{sec:Method}
Our goal is to train a neural network that estimates the 2D and 3D pose of a person, given a single monocular image. Following the analysis-by-synthesis paradigm, we first extract the pose of the person from the input image and then reconstruct the image using this pose data as input. With a naive auto-encoder formulation, given enough freedom, the model can simply output a copy of the input image, without learning anything useful. To prevent that, we use a dual representation of the pose as a vector of keypoint coordinates and a skeleton image, similar to the one introduced in \cite{Jakab:2020:CVPR}. This dual representation creates a tight bottleneck that disentangles pose from appearance, and allows the use of image-to-image translation networks to reconstruct the input image. Separating pose from appearance allows the model to specialize on a single task, improving the performance. 
Since the pose alone lacks appearance information, the image reconstruction is ill posed, and hence, an additional but different image of the same subject is provided as the appearance code.
Furthermore, to ensure the predicted poses are plausible, the distribution of the pose space is constrained using a pose prior, learned from the unpaired pose data through adversarial training.

Overall, our model consists of five components, as depicted in \figref{fig:Overview}. First, the \emph{Skeleton Image Encoder} maps the input image to the skeleton image representation. The \emph{2D Pose Estimator} and \emph{3D Uplift} then predict keypoint coordinates and 3D joint positions as well as orientations. After reprojecting the joint positions, we create the \emph{Analytic Skeleton Image}, from which the \emph{Image Renderer} then reconstructs the input image. 
\FigSingleChannel


\subsection{Multi-Channel Skeleton Image} \label{sec:multichannel}

The skeleton image $y$, introduced in \cite{Jakab:2020:CVPR}, is a pictorial representation of the pose, where each pixel's intensity is computed from the distance to the closest limb:
\begin{equation}
    y = \exp \left( 
        -\gamma 
        \min_{\substack{
            (i, j) \in E \\ 
            t \in[0, 1]}
        } 
        || u - ((1 - t) \cdot p_i + t \cdot p_j) ||^2 
    \right),
    \label{eq:SkeletonImage}
\end{equation}
where $E$ is the set of connected keypoint pairs $(i, j)$, $p$ a keypoint position, $u$ an image pixel coordinate, and $\gamma$ a scaling factor (we used $\gamma = 250$ for our experiments).
To solve the left/right ambiguity of the single-channel skeleton image shown in \figref{fig:SingleChannel}, we extend this representation to a multi-channel image $y \in \mathbb{R}^{C \times W \times H}$, where $C$ is the number of channels.
Each channel represents a semantically meaningful set of joints.
While the exact separation and number of channels may vary, we found empirically that a separation into left and right sides, as well as individual limbs as depicted in \figref{fig:MultiChannel} yielded best performance.
This representation is powerful enough to resolve the ambiguity issues as shown in \figref{fig:MultiChannelResult}, without requiring any changes to the model architecture/capacity nor drastically increase the memory or computation requirements. 

\FigMultiChannel
\FigMultiChannelResult

While our new pose representation should reduce left and right ambiguities, the unsupervised training used in \cite{Jakab:2020:CVPR} lacks conditioning to train successfully. The result is a model capable of generating the input image, without learning keypoints that match the pose in the image.
To solve this problem, we pre-train our model with labelled data that was computer-generated, providing the initial conditioning required for further fine tuning.

\subsection{Uplifting to 3D} \label{sec:uplift}
Prior works using analysis-by-synthesis for pose estimation, only predict 2D poses. And supervised approaches for 3D pose estimation commonly train a separate 2D-to-3D uplift model due to lack of images with 3D annotations. Since such a model is trained from ground truth 2D keypoints, it does not generalize to the noise of a 2D pose estimator, and modelling that noise during the training is hard. Consequently, the 3D pose is not as accurate and does not overlap well with the image. To improve this, we include an uplift module \cite{Martinez17} in the middle of our model (see \figref{fig:Overview}) and train it end-to-end. Specifically, we uplift the predicted 2D keypoints and then reproject the 3D joint positions before creating the analytic skeleton image. For the reprojection, we use a perspective camera. Since we do not know the intrinsic camera parameters, we fix them to plausible defaults (i.e. field of view of 62°). The image reconstruction loss then pushes the model towards 3D poses that overlap with the image, as can be seen in \figref{fig:Uplift}.

\FigUplift

\section{Training} \label{sec:training}
\comment{
Our multi-channel skeleton image extension adds complexity and reduces the constraints in the model. This causes the unsupervised training to be more susceptible to bad local minima and when training from scratch, any of the modules may fail. For example, the predicted pose may look plausible but does not overlap with the image, e.g.\ positions and scale are wrong, despite the learned rendering matching with the input image. The renderer can be powerful enough to compensate for mistakes, but the resulting poses will not be usable.
}
We use a mixture of synthetic and unlabelled real data in two steps. First, we use the synthetic data for pretraining to bootstrap the model, then  train in an unsupervised way on real data, and lastly (as well as optionally) improve the performance for a target video with an unsupervised instance-specific refinement step. 

\paragraph{Pretraining on Synthetic Data}
To bootstrap the model, we pretrain it in a supervised way. With the synthetic data, we can pretrain the individual components independently, which converges faster than training end-to-end. The pretrained components might not work well together, but we found that it is not necessary to refine the model end-to-end on synthetic data. Even though the pretrained model has difficulties to generalize to real data, despite domain randomization, we found that it suffices for bootstrapping the model for the unsupervised training.

\paragraph{Unsupervised Training on Real Data}
With the unlabelled real data, we then continue with the end-to-end training in an unsupervised way. While the image data changes, we reuse the same pose data from before for the pose prior. Besides optimizing several objectives described in \secref{sec:losses}, we continue with synthetic supervision, but with a reduced weight. This helps to regularize and better condition the training, without constraining the model to the synthetic domain. The additional supervision prevents the model from diverging and learning keypoints that do not match the pose, while the input image is successfully rerendered. The trained model then generalizes well to similar real data.

\paragraph{Instance-Specific Refinement}
Given a completely new, unseen video, depending on the similarity to the training data, the model might still struggle e.g. for a different person in a different environment. To address that we use an instance-specific refinement step. To refine the model for a particular person we first collect a few videos of this person, performing various motions. Starting from the trained model, we replace the unlabelled, real training data with the videos of the target person and continue training for a short time. The refined model then generalizes better to videos of the target person.
Alternatively, we can refine on a single target video, by using just this video as the unlabelled training data, which considerably improves the performance, as shown in \figref{fig:InTheWild}. In this case the refinement can also be seen as a post-process optimization step.

\subsection{Training Objectives} \label{sec:losses}
For the unsupervised training to succeed, we optimize several objectives, as shown in \figref{fig:Overview}. This includes losses on the reconstructed image (render loss), losses on the predicted pose (pose prior) as well as losses on synthetic data.

\paragraph{Render Loss}
To train our model, we use a dataset of $N$ images $\{x_i\}_{i=1}^N$ to optimize the reconstruction loss. Instead of directly comparing pixels we use the perceptual loss \cite{NIPS2016_371bce7d}, which compares features extracted from different layers of a pretrained feature extractor $\Gamma$, such as VGG~\cite{Simonyan15}:
\begin{equation}
    L_{perc\_img} = \frac{1}{N} \sum_{i=1}^{N} ||\Gamma_l(x_i) - \Gamma_l(\hat{x}_i)||_2^2~,
\end{equation}
where $\hat{x}_i$ is the reconstructed image and $\Gamma_l$ the extracted features at layer $l$. The losses for the different layers are averaged.
Additionally, we use adversarial training with a multi-scale discriminator $D$ \cite{Wang_2018_CVPR} to capture features at different scales (1, $\frac{1}{2}$, $\frac{1}{4}$).
For the discriminator loss we employ a least square loss \cite{Mao_2017_ICCV}:
\begin{equation} \label{eq:disc}
    L_{disc\_img} = \sum D(x_{real})^2 + \sum (1 - D(x_{fake}))^2.
\end{equation}
Similar to the perceptual loss we also use a discriminator feature matching loss \cite{Park_2019_CVPR}, where the intermediate features of the discriminator are compared:
\begin{equation} \label{eq:disc}
    L_{disc\_img\_FM} = \frac{1}{N} \sum_{i=1}^{N} |D_{l}(x_i) - D_{l}(\hat{x}_i)|~,
\end{equation}
where $D_{l}$ are the features at layer $l$. Both discriminator losses are averaged over the different scales.
\paragraph{Pose Prior}
Besides the unlabelled images, we use a dataset of $M$ unpaired poses from which we create skeleton images $\{\bar{y}\}_{i=1}^M$ with \eqref{eq:SkeletonImage} to build a pose prior to encourage the model to predict plausible poses $y$. Similar to the render loss, we use a multi-scale discriminator $D_{sk}$ for the skeleton images $y$, with a least squares discriminator loss:
\begin{equation} \label{eq:disc}
    L_{disc\_sk} = \sum D_{sk}(y_{real})^2 + \sum (1 - D_{sk}(y_{fake}))^2.
\end{equation}
Additionally, to ensure the estimated 2D and 3D pose follows our desired structure, we have an L2 reconstruction loss $L_{rec\_sk}$ between the predicted skeleton image $y$ and the analytically reconstructed skeleton image $\hat{y}$ from the keypoint 2D coordinates as well as the reprojected 3D positions (for simplicity \figref{fig:Overview} shows only the skeleton image created from the reprojected positions). 

\paragraph{Synthetic Supervision}
During pretraining of the individual modules, we use L2 losses on the predicted skeleton image $y$, the 2D coordinates $p_{2D}$, and the 3D positions and orientations $p_{3D}$. For the reconstructed image, we use the same discriminator losses as explained above. During the training on real data, we keep only the L2 losses shown in \figref{fig:Overview}. The discriminator loss would push the renderer towards synthetic images, which only harms the training.

\paragraph{Overall Learning Objective} Combining all losses and balancing their contributions yields the overall objective (the weights can be found in the supplementary material). Similar to any adversarial formulation the loss on the unlablled data is maximized with respect to the two discriminators and minimized with respect to the other components.

\subsection{Data Augmentation} \label{sec:Augmentations}
The prior work of \cite{Jakab:2020:CVPR} always assumes a tight crop and thus the person will always roughly be at the same scale, which significantly simplifies the problem. But for in-the-wild videos, where no crop is available, such a system will fail. Similarly, if the training data contains only a few different proportions (e.g.\ as in the Human3.6M dataset), the model will have a hard time generalizing to new proportions. This is especially problematic for the 2D-to-3D uplifting, where the predicted 3D pose will be wrongly deformed along the depth axis in order for the reprojection to match.

To generalize better to those cases, we rely on data augmentation. To address the issue with the scale, we randomize the size and offset of the crop. And to address the issue with the proportions, we randomize the limb lengths of the pose data in 3D space and reproject them to get the corresponding 2D coordinates. This way the model can generalize better to different scales and proportions. This also resolves proportion mismatches between the pose prior and the real training data. If we do not randomize the proportions, the model will for example consistently predict a smaller skeleton, but with the right pose. The image renderer then compensates for this scale difference to minimize the render loss. Since there is no overlap for such predictions, they are not usable.


\section{Evaluation} \label{sec:evaluation}
To evaluate the performance of our model, we compare our multi-channel representation to the single-channel representation of \cite{Jakab:2020:CVPR} on the Human3.6M dataset. Further we show qualitative results for in-the-wild videos, as well as other domains such as animals.

\FigHM

\subsection{Human3.6M} \label{sec:EvaluationHM36}
The Human3.6M dataset \cite{h36m:2011,h36m:pami} contains 3.6 million 2D and 3D human pose annotations for 15 different activities, recorded from 4 different viewpoints with static backgrounds. Following the standard protocol, we use subjects 1, 5, 6, 7, and 8 for training and subjects 9 and 11 for evaluation \cite{Villegas_2017}. To compare to prior work, we follow the same scheme as \cite{Jakab:2020:CVPR} and use a disjoint set of the dataset for the pose prior. Because this data has no facial keypoints, we use only a 5-channel skeleton image. While a pose prior from from our in-house motion capture data, containing facial keypoints, also works (as shown with the in-the-wild results), the differences in the skeleton structure prevent a proper comparison to prior work (e.g. offsets in keypoint locations lead to offsets in the error metric).
\figref{fig:HM36} shows qualitatively the accuracy of our 5-channel model on the test set.
To validate the effectiveness of our approach we compare different skeleton image representations (1, 3 and 5 channels), before and after refinement, and the importance of the synthetic supervision. The results summarized in \tabref{tab:Human3.6M:IgnoreFlip} and \tabref{tab:Human3.6M:WithFlip} show that our approach outperforms the single-channel baseline \cite{Jakab:2020:CVPR} and significantly reduces flips.
After the unsupervised training on real data we achieve similar or better scores for almost all activities and after refining we outperform the baseline by a large margin. 

\TableIgnoreFlip
\TableConsiderFlip


\subsection{In-the-Wild Videos} \label{sec:EvaluationInTheWild}
Whereas for annotated datasets we can crop around the person for optimal performance, for in-the-wild footage this is not available. Since we used in-the-wild videos for our real dataset, combined with data augmentation, our model is more robust to such cases. As can be seen in \figref{fig:InTheWild} and our accompanying video, the estimated poses for people used during training are very accurate. While for a completely different person the model might have some difficulties, it can quickly adapt with the instance-specific refinement.


\subsection{Animals} \label{sec:EvaluationAnimals}
Our framework is not specific to humans, but can also work for other skeleton structures, like dogs or lions. For wild animals it is even more inherent that capturing the motion with traditional motion capture setups is difficult.
Other works such as \cite{Zuffi:ICCV:2019,Borer21:CatsAndDogs} only used synthetic data to generalize to wild animals, but this remains challenging due to the reality gap and in many cases the estimated 3D poses do not overlap well with the image.
With our approach on the other hand we can leverage real footage of wild animals and the estimated 3D poses are more accurate and have a better overlap due to the end-to-end training, as shown in \figref{fig:Lion} and our accompanying video. 
Furthermore, with the instance-specific refinement we can easily adapt to other breeds e.g.\ a model trained only on lions can be refined on a video of a dog.

\FigInTheWild

\FigAnimals
 

\section{Limitations and Future Work}
\comment{
While we are able to train a 2D and 3D pose estimator from unlabelled data without suffering from the left/right ambiguity issue from the previous work, we still require annotated synthetic data for bootstrapping the training and for regularization. Being able to start from scratch in an unsupervised way, without losing performance, would therefore be very beneficial.
}

One of the limitations of our model is the resolution of the rendering module ($128\times128$ pixels). Increasing this resolution and quality of the rendered image, while challenging, could help further increase the effectiveness of the analysis-by-synthesis approach. Additionally, it would also be interesting to explore other image modalities such as UV, normal and depth maps and integrate them into the pipeline to help with the pose estimation as well as with the rendering.

Although we extended the pose representation to multiple channels, we are challenged to extend further, say to one channel per bone.  We believe future work with a tailored architecture and training curriculum could make this feasible.

Besides the use case of monocular pose estimation, it would be interesting to explore the effectiveness of analysis-by-synthesis in a multi-camera scenario, where we could make use of multi-view consistency. A multi-camera setup is often available in lab scenarios, for example to study the behaviour of animals as in \cite{OpenMonkeyStudio:2020}. But currently employed techniques still rely on large, manually annotated datasets of the animal to study, which is expensive to create.

\section{Conclusion}
We have shown that by extending the skeleton image representation to multiple channels in conjunction with mixing synthetic and unlabelled real data, we can reduce left and right ambiguity issues, avoid bad local minima and address the reality gap, without having to manually label data. Furthermore, we extended the analysis-by-synthesis pose estimation framework to predict 3D poses that overlap well with the input image thanks to the end-to-end training. And lastly, we presented an instance-specific refinement that allows us to considerably increase the performance on a completely new, unseen target video, without requiring any additional annotations for the target subject or environment.

{\small
\bibliographystyle{unsrtnat}
\bibliography{analysis-by-synthesis}
}

\end{document}